\documentclass{ceurart}
\hypersetup{breaklinks=true}
\def\ONT{\textsc{OASIS}}

\begin{document}
\copyrightyear{2023}
\copyrightclause{Copyright for this paper by its authors.
Use permitted under Creative Commons License Attribution 4.0
International (CC BY 4.0).}

\conference{7th Workshop on Foundational Ontology (FOUST) co-located with FOIS 2023, 19-20 July, 2023, Sherbrooke, Québec, Canada. Corresponding author: D. F. Santamaria.}

\title{A behaviouristic approach to representing processes and procedures in  the OASIS 2 ontology}

\author[1]{Giampaolo Bella}[%
email=giampaolo.bella@unict.it,
url=https://www.dmi.unict.it/giamp/,
]
\address[1]{Department of Mathematics and Computer Science, University of Catania, Viale Andrea Doria 6 - 95125 - Catania, Italy}

\author[1]{Gianpietro Castiglione}[%
email=gianpietro.castiglione@phd.unict.it
]

\author[1]{Daniele Francesco Santamaria}[%
email=daniele.santamaria@unict.it,
url=https://www.dmi.unict.it/santamaria/
]

\begin{abstract} 
Foundational ontologies devoted to the effective representation of processes and procedures are not widely investigated at present, thereby limiting the practical adoption of semantic approaches in real scenarios where the precise instructions to follow must be considered. Also, the representation ought to include how agents should carry out the actions associated with the process, whether or not agents are able to perform those actions, the possible roles played as well as the related events. 

The \ONT{} 2 ontology~\cite{ia2022, oasis2} provides an established model to capture agents and their interactions but lacks means for representing processes and procedures carried out by agents. This motivates the research presented in this article, which delivers an extension of the \ONT{} 2 ontology to combine the capabilities for representing agents and their behaviours with the full conceptualization of processes and procedures. The overarching goal is to deliver a foundational OWL ontology that deals with agent planning, reaching a balance between generality and applicability, which is known to be an open challenge.
 
\end{abstract}


\maketitle

\section{Introduction}

The notions of process and procedure are broadly known in the literature, even outside the computer science field, although they bear subtle differences. According to ISO 9001-2015 definitions, a process is \emph{a set of correlated or interactive activities that transform inputs into outputs} describing \emph{the specified way of carrying out an activity}, whereas a procedure describes \emph{how to carry out all or part of a process}. Hence, we can argue that a process may be composed of many procedures.
 
Representing processes and procedures through foundational \emph{Web Ontology Language} (OWL) ontologies is necessary both for easing the exchange of process descriptions among systems and to provide a machine-understandable interpretation of the related activities. One of the main benefits of using Semantic Web technologies in such a context is the reasoning capability that permits the inference of new facts from existing data and the verification of the knowledge base. For example, in the context of operating systems, \emph{race conditions} \cite{netzer92} could lead to problems related to the lock of resources by concurrent processes. A formalization of the processes, procedures and agents together with semantic reasoning could reduce the ambiguity that may occur when agents attempt to perform more operations at the same time.

Defining an ontological architecture for processes and procedures aims at improving  so-called ``automated planning and acting'' \cite{ghallab_nau_traverso_2016}. One of the most remarkable attempts to standardize the planning problem is the PDDL (Planning Domain Definition Language) \cite{pddl}, which separates the planning problem into a) domain description and b) problem description. For instance, the PDDL in its current version introduces \emph{derived predicates} for dependency modelling of properties between objects. Therefore, processes and procedures benefit from the adoption of ontological representations because ontologies can fully address both planning sub-problems.  
 
Although many ontological approaches are available in the literature, they suffer from the lack of a complete and general approach to effectively represent processes and procedures, especially one that a)  combines the representation of processes and procedures with agents and their commitments, b) models the events generated during the executions of processes and c) accounts for the roles that are played by the committer agents. 

The current paper is motivated by the delivery of such a model. We start by considering the ontological foundations for the domain of multi-agent systems. Specifically, we take into account \ONT{} 2~\cite{ia2022, oasis2}, a foundational OWL 2 ontology that leverages the behaviouristic approach derived from the \emph{Theory of Agents} and the related mentalistic notions. The behaviouristic approach is an effective way of semantically describing agents by characterizing their capabilities. Agents are enabled to report the set of activities that they can perform, the types of data required to execute those activities as well as the expected outputs, through the description of their behaviours. Agents' implementation details are abstracted away to make the discovery of agents transparent, automatic, and independent of the underlying technical infrastructure. In consequence, agent commitments are clearly described, and the entire evolution of its environment is unambiguously represented, searchable, and accessible:  agents may join a collaborative environment in a plug-and-play fashion, as there is no more need for third-party interventions. 

Our work rests on the observation that \ONT{} 2 has lacked a specific characterization of processes and procedures so far, even though it models so-called \emph{plans}, which are ways of depicting a sequence of (planned) actions to be tackled once. Therefore, plans cannot be applied again once they are performed by agents. In this paper, we extend \ONT{} 2 to deal with general specifications of processes and procedures that can be consumed by agents, including events and the played role, that can be practically leveraged in real scenarios such as the one concerning race conditions.

The paper is organized as follows. Section \ref{sec:related} presents the related work through a comparison with the \ONT{} 2 approach; Section \ref{sec:prelim} introduces the model of \ONT{} 2 devoted to the representation of agents and their behaviours. Section \ref{sec:process} presents the novel extension of \ONT{} 2 that deals with the notions of processes and procedures. Section \ref{ref:conclusions} closes the paper with some hints for future outcomes.

\section{Related Work} \label{sec:related}
In \textsf{DOLCE} \cite{gangemi2002dolce}, the concepts of processes and events are presented as special types of perdurants, whereas functions and roles are formalized in some extensions of the ontology \cite{masolo04, borgo10}.  Aware of the general definitions provided by DOLCE, \ONT{} 2 focuses on the behaviouristic approach which is leveraged to provide agents with novel means for representing complex planning and related actions, dealing also with events. Specifically, roles are conceived by \ONT{} 2 as ways of enabling agents with additional behavioural capabilities, whereas events are represented to be aligned with the definition of agent behaviours. From this point of view, \ONT{} 2 introduces a different conceiving of events and roles since it describes agents in terms of their capabilities. A full mapping of \ONT{} 2 in DOLCE is feasible and one of our planned future works.

Wang et al. introduce a model for processes related to water quality monitoring \cite{processOntology2020}. The model is strictly focused on the description of observational processes concerning water pollution monitoring, thus being limited to general applicability outside of the domain.
 
Within CIDOC-CRM \cite{Bruseker2017}, there is a work-in-progress extension called CRMact \cite{CRMact} that defines the classes and properties for planning future activities and events. However, CRMact is mainly focused on cultural heritage and documentation records and hence is not generally applicable.

Concerning business processes, it is worth mentioning several works. Thomas et al. propose an ontological approach for representing business processes together with a system architecture prototype, exploiting the proposed model \cite{Thomas2009SemanticPM}. Greco et al. use ontologies as facilities within a framework for assisting designers in the realization and analysis of complex processes \cite{greco04}. Corea et al. present an approach to verify whether a business process is compliant with given business rules combining logic programming and ontologies \cite{Corea2017}, while Calvanese et al. propose an approach for using Ontology-Based Data Access (OBDA) to govern data-aware processes, and in particular, those executed over a relational database that issue calls to external services to acquire new information and update data \cite{calvanese12}. 
Finally, a comparison among ontologies related to business processes for task monitoring, measurements and evaluation strategies, and for modelling process information was built \cite{oatao26340}, \cite{BECKER2015}, \cite{ISARC2015}.

However, the downside of limiting ontological models to business processes lies in the absence of relationships between agents and their commitments that instead represent the core of agent-oriented representation. That implies the inability of finding agents/services with specific capabilities, invoking them, and enabling their interoperability. Notably, the benefits of process-oriented representations, such as the facilitation mechanisms for the search and selection of process models, are also offered by the agent-oriented ones as long as they are sufficiently general although flexible.

\section{Preliminaries on \ONT{} 2} \label{sec:prelim}
The first version of \ONT{} \cite{woa2019} is a foundational ontology that leverages the behaviouristic approach to characterize agents in terms of the actions they are able to perform, including purposes, goals, responsibilities, information about the world they observe and maintain, and their internal and external interactions. It models the executions and assignments of tasks, restrictions and constraints used to establish agent responsibilities and authorizations.

In the recent past, \ONT{} has been extended and applied to deal with  so-called \emph{Ontological Smart Contracts}~\cite{idc2021} and with the ontological models for smart contracts on the blockchain~\cite{idc2021-2}. \ONT{} is also part of the POC4COMMERCE project \cite{gecon21}, funded by the NGI-ONTOCHAIN consortium \cite{ontochain}.

The last version of \ONT{} is \ONT{} 2 \cite{woa2022, ia2022}\footnote{The \ONT{} 2 ontology can be reached at \cite{oasis2}}, which extends \ONT{} with some new features, such as the entrustment of agents, and reshapes the model adopted for the representation of agents and their commitments.

Inspired by the \emph{Tropos} methodology \cite{tropos} which devises from Agent Oriented Programming (AOP), \ONT{} 2 represents agents through three essential and publicly shared mental states, namely (expected) \emph{behaviours}, \emph{goals} and \emph{tasks}. Behaviours represent the mental state of the agent associated with its ability to modify its environment or, in general, act or do something. Goals describe mental attitudes representing preferred progressions of a particular system that the agent has chosen to put effort into bringing about \cite{riemsdijk08}. Tasks depict how to carry on such progressions and describe atomic operations that agents perform.  

Agents and their interactions are represented by carrying out three main steps, namely: a) an optional step that consists of modelling descriptions of general abstract behaviours, called \emph{templates}, conceptual characterization of behaviours from which concrete agent behaviours are drawn; b) modelling concrete agent behaviours, possibly, drawn by agent templates; c) modelling actions and associating them with the corresponding behaviours.
The first step, not mandatory, consists in defining the agent's behaviour template, namely a higher-level description of the behaviour of abstract agents that can be leveraged to define concrete behaviours of real agents; for example, a template is designed to describe the abstract behaviour consisting in obtaining and releasing locks on resources. Additionally, templates are useful to guide developers in the definition of the behaviours of their specific agents. To describe abstract agent's capabilities to perform actions, an agent template comprises three main elements, namely behaviour, goal and task. The latter constitutes the most simple (atomic) operations that agents are able to actually perform including, possibly, input and output parameters required to accomplish them.
The second step consists of representing concrete agent behaviours either by relying on a template or by defining it from scratch. In both cases, concrete behaviours are modelled analogously to those of templates, where the models of outstanding features are replaced with actual characteristics. Behaviours drawn by shared templates are associated with them in order to depict the behaviour inheritance relationship.
In the last step, actions performed by agents are described as direct consequences of some behaviours and are associated with the behaviours of the agent that performed them. To describe such an association, \ONT{} 2 introduces \emph{plan executions}. Plan executions describe the actions performed by an agent, associating them with one of its behaviours. Associations are carried out by connecting the description of the performed action to the behaviour from which the action has been drawn: actions are hence described by suitable graphs that retrace the model of the agent's behaviour. 

Plans can be additionally either submitted to agents as requests for performing some actions or they can be assigned by specific agents called \emph{entruster agents}.

\begin{figure*}[ht]
    \includegraphics[scale=0.74]{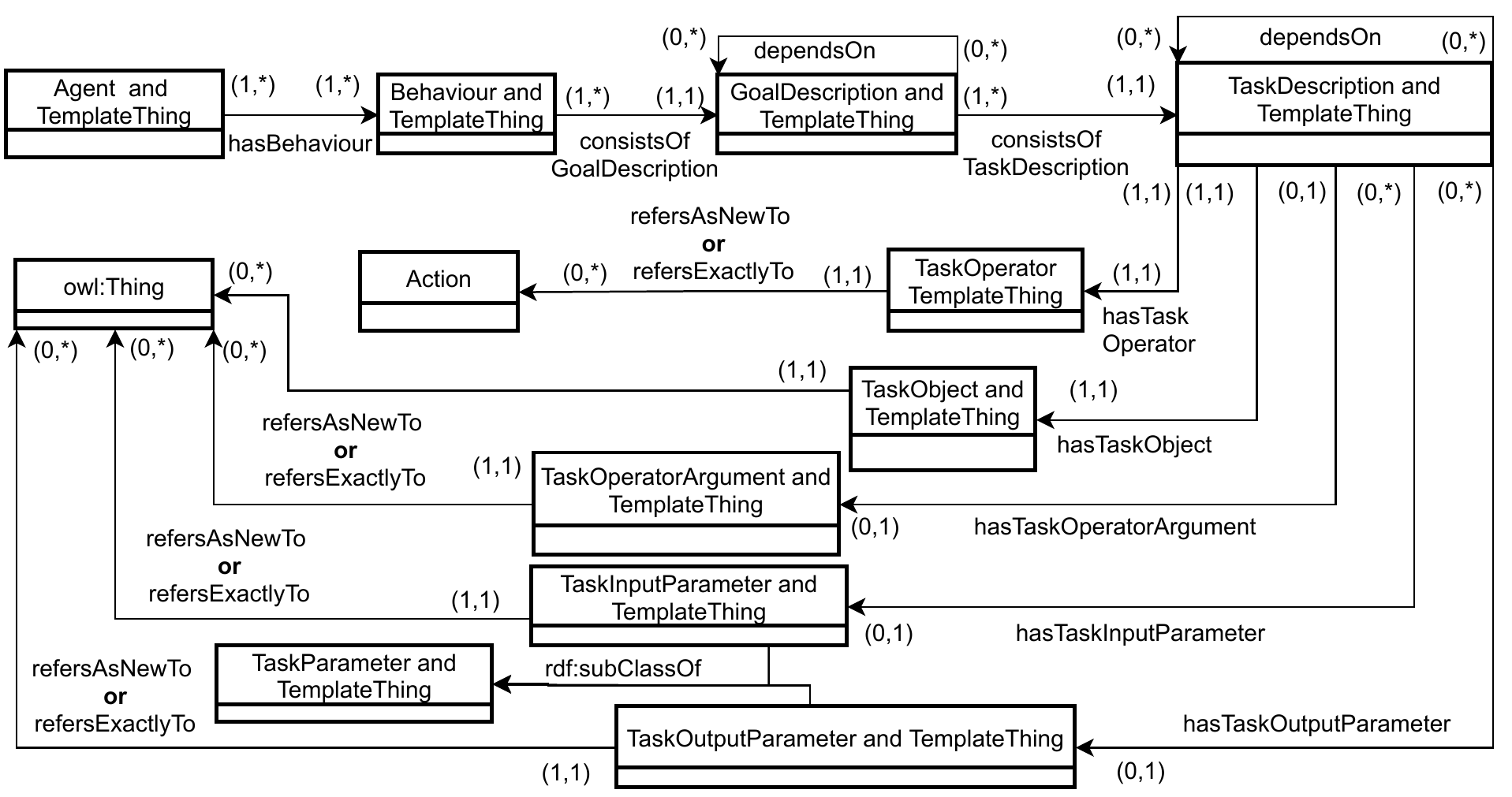}
    \caption{Diagram of agent templates in \ONT{} 2}
    \label{fig:oasis:agent-template}
\end{figure*}

In \ONT{} 2, agent templates are defined according to the UML class diagram in Fig. \ref{fig:oasis:agent-template}. 
To consider how both abstract and concrete agents perform actions, the description of agents comprises three main elements, namely \emph{behaviour}, \emph{goal}, and \emph{task}. Agent tasks, in their turn, describe atomic operations that agents perform, including possibly input and output parameters required to accomplish them. Those elements in \ONT{} 2 are introduced by way of the following OWL classes:

\begin{itemize} 
    \item \textit{Agent}: This class comprises all the individuals representing agents. Instances of such a class are connected with one or more instances of the class \textit{Behaviour} using the OWL object-property \textit{hasBehaviour}.
    
    \item \textit{Behaviour}: Behaviours can be seen as collectors comprising all the goals that an agent may achieve. Instances of \textit{Behaviour} are connected with one or more instances of the class \textit{GoalDescription} by means of the object-property \textit{consistsOfGoalDescription}.
    
    \item \textit{GoalDescription}: Goals represent containers embedding all the tasks that the agent can achieve. Instances of \textit{GoalDescription} comprised by a behaviour may also satisfy dependency relationships introduced by the object-property \textit{dependsOn}. Goals are connected with the tasks that form them and are represented by instances of the class \textit{TaskDescription} through the object-property \textit{consistsOfTaskDescription}.
    
    \begin{sloppypar}
    \item \textit{TaskDescription}: This class describes atomic operations that agents perform. Atomic operations are the most simple actions that agents are able to execute and, hence, they represent what agents can do within their environment. Atomic operations may depend on other atomic operations when the object-property \textit{dependsOn} is specified. Atomic operations whose dependencies are not explicitly expressed are intended to be performed in any order. 
    \end{sloppypar}
\end{itemize}

The core of agent behaviour revolves around  the description of atomic operations represented by instances of the class \textit{TaskDescription} that characterizes the mental state corresponding to commitments. In their turn, instances of the class \textit{TaskDescription} are related to the following five elements that identify the operation: 

\begin{itemize}
    \item An instance of the class \textit{TaskOperator}, characterizing the mental state corresponding to the action to be performed. Instances of \textit{TaskOperator} are connected either by means of the object-property \textit{refersExactlyTo} or \textit{refersAsNewTo} to instances of the class \textit{Action}. The latter class describes physical actions represented by means of entity names in the form of infinite verbs (e.g., \textit{produce}, \textit{sell}). Specifically, the object-property \textit{refersExactlyTo} is used to connect the task operator with a precise action having a specific IRI,  whereas \textit{refersAsNewTo} is used to connect a task operator with an entity  representing an action of which only a general abstract  description is given (for example, an action for which only the type is known). 
    
    In the latter case, the entity representing the action is also defined as an instance of the \textit{TemplateThing}: such instances are used to define entities that represent templates for the referred element and that describe the characteristics that such element should satisfy. \textit{TemplateThing} is the class used to characterize all the individuals involved in the definition of  behaviour templates and to distinguish them from the entities representing concrete behaviours, plans or actions, thus eliminating the need of having separated models for those aspects.
    
    In order to specify the classes of which the entity must be an instance, it is eventually possible to connect such entity by means of the object-property \textit{refersAsInstanceOf} with the individual instances of the desired classes.

    \item Possibly, an instance of the class \textit{TaskOperatorArgument}, connected using the object-property \textit{hasTaskOperatorArgument} and representing additional specifications or subordinate characteristics of  task operators  (e.g., \textit{on}, \textit{off}, \textit{left}, \textit{right}). Instances of \textit{TaskOperatorArgument} are referred to the operator argument by using either the object-property \textit{refersAsNewTo} or \textit{refersExactlyTo}. 
    
    \item An instance of the class \textit{TaskObject}, connected by means of the object-property \textit{hasTaskObject} and representing the template of the object recipient of the action performed by the agent (e.g., \textit{price}). Instances of \textit{TaskObject} are referred to the action recipient by specifying either the object-property \textit{refersAsNewTo} or \textit{refersExactlyTo}. 
           
    \item Input parameters and output parameters are introduced by instances of the classes \textit{TaskInputParameter} and \textit{TaskOutputParameter}, respectively. Instances of  \textit{TaskDescription} are related to instances of the classes  \textit{TaskInputParameter} and \textit{TaskOutputParameter} by means of the object-properties \textit{hasTaskInputParameter} and \textit{hasTaskOutputParameter}, respectively. Instances of \textit{TaskInputParameter} and  of \textit{TaskOutputParameter} are referred to the parameter by specifying either the object-property \textit{refersAsNewTo} or \textit{refersExactlyTo}. Moreover,  the classes \textit{TaskInputParameter} and \textit{TaskOutputParameter} are also subclasses of  \textit{TaskParameter}. 
    
\end{itemize}

\begin{sloppypar}
Finally, in the case of agent behaviour templates,  instances of \textit{Agent}, \textit{Behaviour}, \textit{GoalDescription}, \textit{TaskDescription}, \textit{TaskOperator}, \textit{TaskOperatorArgument}, \textit{TaskObject} \textit{TaskInputParameter}, and \textit{TaskOutputParameter} are also instances of \textit{TemplateThing}. 
\end{sloppypar}

\begin{figure}[ht]
   \includegraphics[scale=0.76]{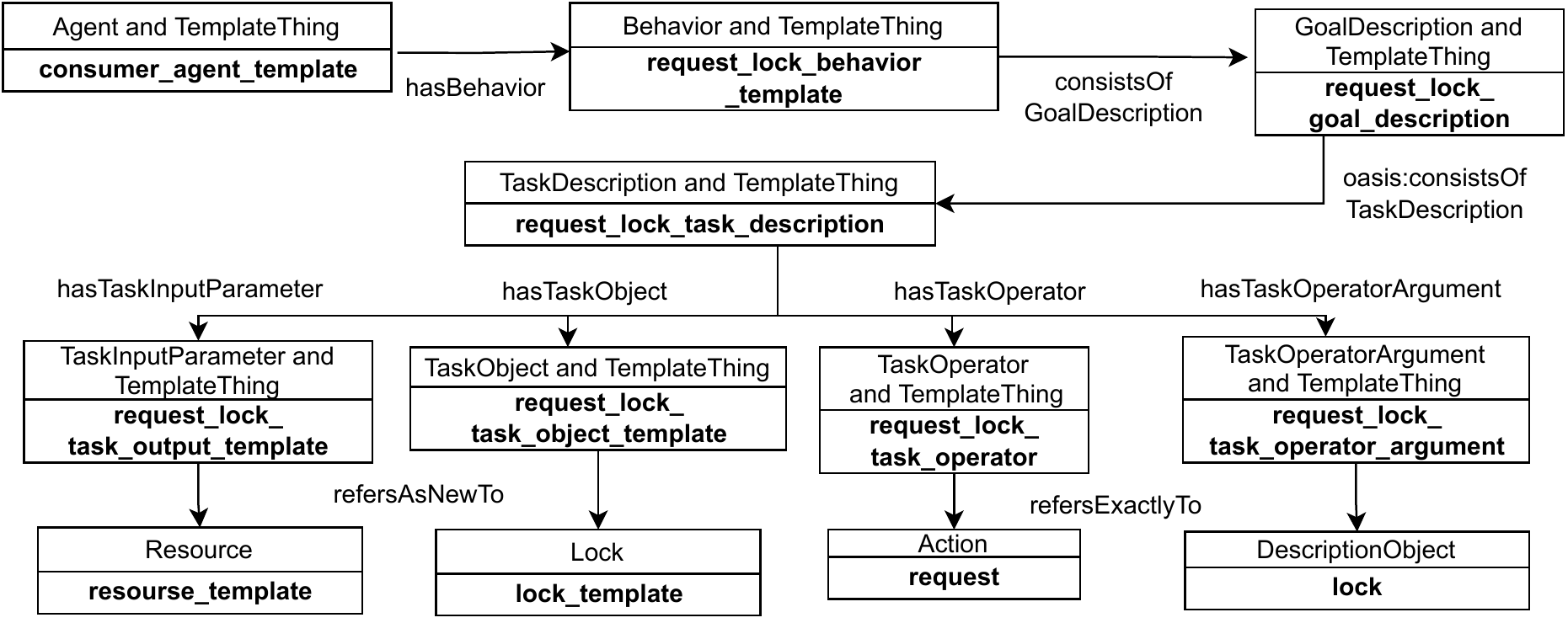}
   \caption{Example of an \ONT{} 2 agent template}
    \label{fig:oasis:agent-template-example}
\end{figure}

Fig. \ref{fig:oasis:agent-template-example} illustrates the case study mentioned above where a lock on a resource can be obtained and released by agents. Specifically, Fig. \ref{fig:oasis:agent-template-example} presents a template describing  an abstract agent that is able to request a lock on a resource. The agent template comprises a single behaviour, constituted by a single goal that in its turn comprises a single task. The task, which represents the ability to request a lock,  provides four elements:
\begin{itemize}
\item  \textit{request\_lock\_task\_operator}, representing the mental state associated with the behaviour's action (the task operator), which in its turn is associated with the individual \textit{request}, the latter describing the capability of requesting something. 

\item \textit{request\_lock\_task\_operator\_argument}, introducing an additional feature (the operator argument) associated with the action and represented by the individual \textit{lock}. The argument describes the fact that the request action is referred to a lock. Task operator and its argument describe together the capability of requesting a lock; 

\item \textit{request\_lock\_task\_object\_template}, representing the recipient of the operation, which is related to an instance of the class \textit{Lock} by means of the object-property \textit{refersAsNewTo}. Such an instance comprises all the features that the recipient of the adopting action should own: the concrete actions implementing the behaviour template for requesting a lock are supposed to effectively request a lock with the desired features;  

\item \emph{request\_lock\_task\_output\_template}, representing the input of the operation, namely a resource on which the lock is requested.
\end{itemize}

In the second step, concrete agent behaviours are defined either by instantiating one or more templates or from scratch. In \ONT{} 2, the modelling pattern of concrete behaviours has a structure analogous to one of the behaviour templates, illustrated above, with the difference that individuals used to define a concrete behaviour, instead of being instances of the class \textit{TemplateThing}, are instances of the class \textit{BehaviourThing}. The latter class is  devoted to describe all the mental states associated with concrete behaviours of real agents that induce actions.

Concrete behaviours may be connected with the template they are drawn from. In order to describe the fact that concrete agents inherit their behaviours from a commonly shared template, the  instances related to the concrete behaviours are connected with the instances of the template through the sub-properties of the object-property \textit{overloads} as follows. The association is carried out by connecting the instances of the classes:

\begin{itemize}
    \item \textit{Behaviour}, by means of \textit{overloadsBehaviour}; 
    \item \textit{GoalDescription}, by means of \textit{overloadsGoalDescription}; 
    \item \textit{TaskDescription}, by means of \textit{overloadsTaskDescription};  
    \item \textit{TaskObject}, by means of  \textit{overloadsTaskObject}; 
    \item \textit{TaskOperator}, by means of \textit{overloadsTaskOperator}; 
    \item \textit{TaskInputParameter}, by means of \textit{overloadsTaskInputParameter};  
    \item \textit{TaskOutputParameter}, by means of the object-property \textit{overloadsTaskOutputParameter}.
\end{itemize}

As the last step, agent commitments devised from behaviours are introduced to describe agent actions. In \ONT{} 2, commitments are represented by adopting the same pattern presented for  abstract behaviours  with the difference that instances of the class \textit{TemplateThing} are instead modelled as instances of the class \textit{ExecutionThing} and the agent responsible for the execution of the action is related with the plan representing the commitment by means of the object-property \textit{performsPlanExecution}, subproperty of \textit{performs}. 
The class \textit{ExecutionThing} is introduced to characterize all the entities involved in the definition of concrete and already performed actions and to distinguish them from the ones introduced for templates, behaviours and plans.

In order to relate agent commitments with the behaviour from which they are drawn, subproperties of the object-property \textit{drawnBy} are introduced. Specifically,  \textit{planExecutionDrawnBy} connects the instance of \textit{GoalDescription} of the agent action to its analogue of agent behaviour; much in the same way, \textit{goalExecutionDrawnBy} connects the instance of the class \textit{GoalDescription} of the commitment with its analogue, while \textit{taskExecutionDrawnBy}, \textit{taskObjectDrawnBy}, \textit{taskOperatorDrawnBy}, \textit{taskInputParameterDrawnBy}, and \textit{taskOutputParameterDrawnBy} are introduced for \textit{TaskDescription}, \textit{TaskObject},  \textit{TaskOperator}, \textit{TaskInputParameter}, and \textit{TaskOutputParameter}, respectively.

\begin{sloppypar}
Usually, agents proposing plans identify the behaviours responsible for their realization beforehand, in such a way as to completely describe and trace how agent intentions are realized. In this case, the entities representing the submitted plan are related to the entities describing the responsible behaviour by means of suitable subproperties of the object-property \textit{submittedTo}, relating instances of \textit{PlanningThing} with instances of \textit{BehaviourThing} as follows: a) \textit{planDescriptionSubmittedTo}, for instances of \textit{Behaviour}; b) \textit{goalDescriptionSubmittedTo}, for instances of \textit{GoalDescription}; 
c) \textit{taskDescriptionSubmittedTo}, for instances of \textit{TaskDescription};  d) \textit{taskObjectSubmittedTo}, for instances of \textit{TaskObject}; e) \textit{taskOperatorSubmittedTo}, for instances of \textit{TaskOperator}; f) \textit{taskInputParameterSubmittedTo}, for instances of \textit{TaskInputParameter}; g) \textit{taskOutputParameterSubmittedTo}, for instances of \textit{TaskOutputParameter}.
\end{sloppypar}

In a similar way, plans are also related to the agent's actions realizing them. For this purpose, the subproperties of the object-property \textit{hasExecution} are introduced, namely \textit{hasPlanExecution}, \textit{hasGoalExecution}, \textit{hasTaskExecution}, \textit{hasTaskObjectExecution}, \textit{hasTaskOperatorExecution}, \textit{hasTaskInputParameterExecution}, and \textit{hasTaskOutputParameterExecution}. Analogously, actions are connected with the behaviour responsible for their execution by means of suitable subproperties of the object-property \textit{executionDrawnBy}.

\section{Processes and procedures in \ONT{} 2} \label{sec:process}
\begin{figure}[ht]
\centering
   \includegraphics[scale=0.80]{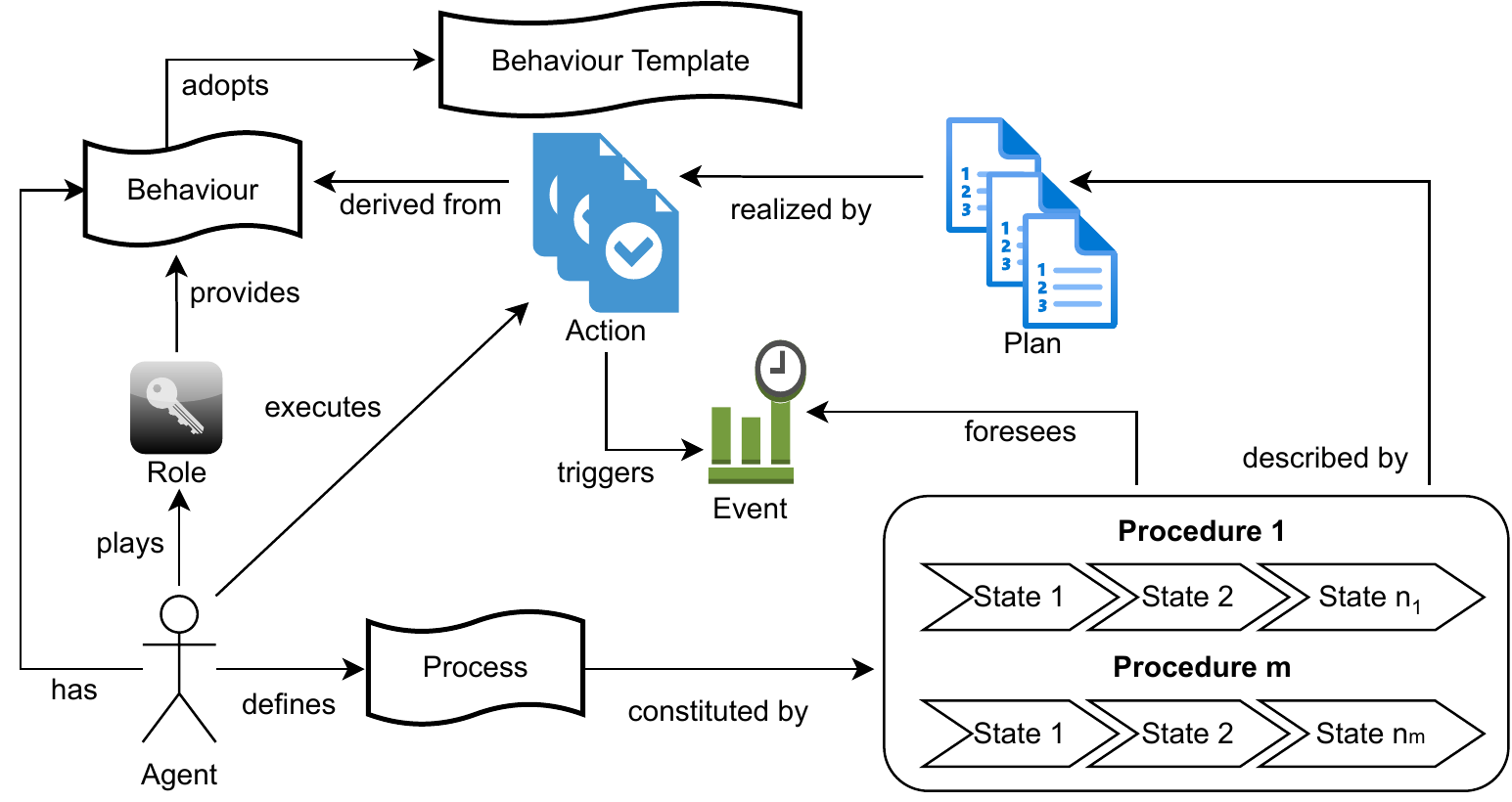}
   \caption{Process and procedure in \ONT{} 2}
    \label{fig:general}
\end{figure}

Since the first version, \ONT{} was already capable of describing how agent activities are carried out by suitably combining behaviours, plans and actions. Plans, in particular, can be used to describe in detail how inputs are processed to be turned into outputs, but they are not sufficient to describe general complex processes. Plans require the presence of a committer agent beforehand and can be applied only once. This is because the mental state related to the desire or wish to perform actions does not abstract from the committer agent. Plans can be leveraged to represent single applications of procedures but they should be combined properly. Then, plans can be associated with the behaviours responsible for their executions thus carrying out the actions required to accomplish the procedures. 
The approach is illustrated in Fig. \ref{fig:general}. Leveraging the ISO definition, processes defined by agents are constituted by procedures. In the behaviouristic vision, the procedures states are associated with plans that are executed thus leading to actions. Actions derive from agent behaviours that in their turn can rely on specific templates. Agents performing actions are provided with behaviours either intrinsically or through a played role. Finally, actions could lead to events either incidentally or as foreseen by procedures. In the first case, events are conceived as noteworthy and extraordinary happenings that have not been previously planned, in the second case, as recurrent situations.

For example, agents may play the role of resource manager, hence they are able to request and release locks on resources according to such role. A process describing the steps required to request and release locks is formalized, so that they can tackle a plan for each step. Plans are executed so that the resource is released and modified. Actions are performed thanks to agent behaviours provided by the resource manager's role. During the execution of one of those actions, the system exceptionally sends a message to the committer agent. We will partially see this scenario together with the presentation of the model.

In light of the above considerations, \ONT{} 2 introduces the following novel OWL classes:

\begin{itemize}
    \item \textit{Process}, which encompasses the procedures describing how activities are carried out and;
    \item \textit{Procedure}, the subclass of the class \textit{Activity}, which introduces the set of plans required to accomplish the procedure itself and to be realized through actions. In \ONT{} 2 procedures are constituted  by one or more \textit{ProcedureState}, each one  connected to a specific plan describing how the activity is carried out. Specifically, we identify two types of \textit{ProcedureState}: a) \textit{TerminatingProcedureState}, which includes \textit{InitialProcedureState} and \textit{FinalProcedureState},  and b) \textit{NonTerminatingProcedureState}.    
\end{itemize}

\begin{figure}[ht]
   \includegraphics[scale=0.70]{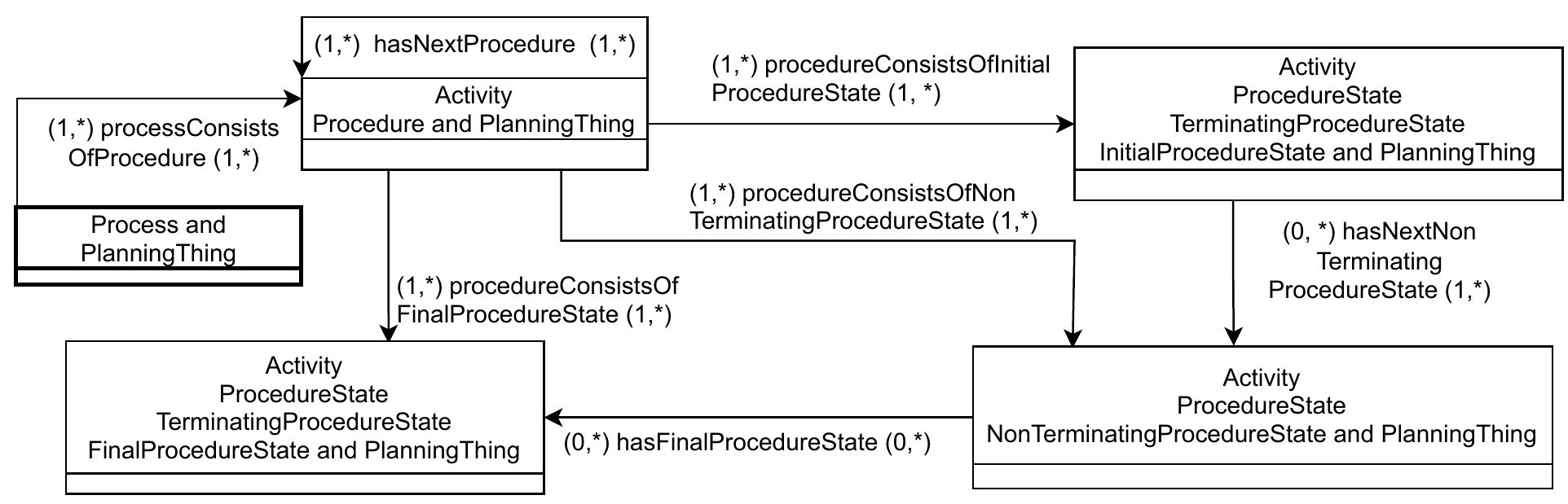}
   \caption{The \ONT{} 2 model for processes and procedures}
    \label{fig:process-schema}
\end{figure}

Procedures are constituted by \emph{procedure states} that describe single steps of the procedure. \emph{Initial procedure states} describe the beginning of the procedure,  while \emph{final procedure states} its termination. Moreover, the initial procedure state coincides with the final procedure state in the case of a single-step procedure. Finally, \emph{non-terminating procedure states} describe the intermediate steps to be performed, including all the steps between the initial state and the final state. 
The schema for processes and procedures is illustrated in Fig. \ref{fig:process-schema}. The subproperties of the object-property \textit{procedureConsistsOfProcedureState} are used to suitably connect a procedure with its procedure states. Specifically, the object-properties \textit{procedureConsistsOfInitialProcedureState} and \textit{procedureConsistsOfFinalProcedureState} (both subproperty of \textit{procedureConsistsOfTerminatingProcedureState}) connect the procedure with its initial and final procedure state, respectively, while the object-property \textit{procedureConsistsOfNonTerminatingProcedureState} connects the procedure with its non-terminating states. The initial procedure state is connected with the subsequent non-terminating procedure states by means of the object-property \textit{hasNextNonTerminatingProcedureState}. The latter property is also used to connect non-terminating procedure states with their subsequent non-terminating procedure states. Whenever the next procedure state is the final state, the object-property \textit{hasFinalProcedureState} is adopted in its place. Both \textit{hasNextNonTerminatingProcedureState} and \textit{hasFinalProcedureState} are defined as subproperty of the object-property \textit{hasNext}.

Finally, since the intended meaning of processes and related procedures is to give a description of how activities should be carried out, the instances introduced so far are also instances of the class \textit{PlanningThing}. To complete the representation of processes, it is now sufficient to connect an instance of the class \textit{Process} with the ones of the class \textit{Procedure} that model the process activities. In case a process is constituted by more than one procedure, the latter can be sorted by connecting them through the object-property \textit{hasNextProcedure}. 

To describe how process activities should be carried out, we introduce a behaviour as in Fig.~\ref{fig:oasis:agent-template}, where the instances of \textit{TemplateThing} are instead defined as instances of \textit{PlanningThing}. The choice is motivated by the fact that templates describe abstract behaviours that are instantiated to introduce concrete behaviours, whereas plans describe actions that agents wish to perform or see accomplished: this is exactly the case of process, where a set of actions must be tackled in order to achieve the desired end. Additionally, plans allow one to describe in detail what actions are required to accomplish the activity associated with the procedure state, the input parameters and the expected output, without the need of selecting an agent responsible for those actions beforehand. To associate a procedure state with the planned behaviour, \ONT{} 2 introduces the object-property \textit{isDescribedBy}.

\begin{figure}[ht]
   \centering
   \includegraphics[scale=0.87]{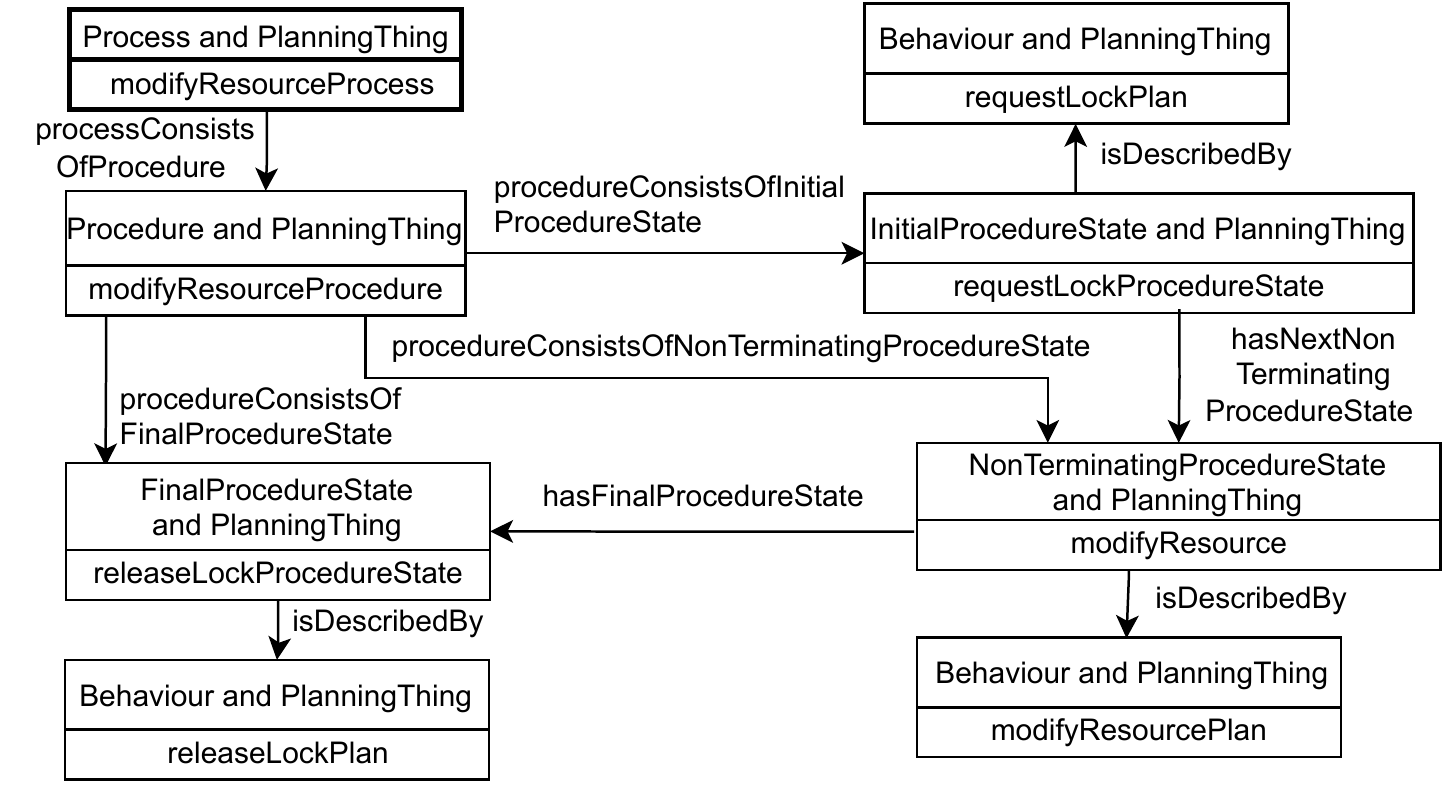}
   \caption{Example of process in \ONT{} 2}
    \label{fig:process-example}
\end{figure}

The process concerning the case study considered at the beginning of the section is depicted in Fig. \ref{fig:process-example}. The process describes the modification of a resource which requires the acquisition and release of an exclusive lock before and after the modification, respectively.

The process consists of a single procedure that, in its turn, consists of three distinct states, an initial state, a non-terminating state, and a final state. The initial state is associated with a behaviour describing how to request a lock on a resource. The behaviour follows the template in Fig. \ref{fig:oasis:agent-template-example}, where instances of the class \textit{TemplateThing} are replaced with instances of the class \textit{PlanningThing} as stated in Section \ref{sec:prelim}. Analogously, the non-terminating state is associated with a behaviour describing how to modify the resource, while the final state is associated with the behaviour describing how to release the lock.

Once the process is fully described, it is possible to model its application. To do so, a model retracing the structure of a process (see Fig. \ref{fig:process-schema}) is introduced, where instances of the class \textit{PlanningThing} are instead defined as instances of the class \textit{ExecutionThing}. In a similar way to \textit{TemplateThing} and \textit{PlanningThing} that are introduced to represent descriptions of abstract behaviours and planned actions, respectively, \textit{ExecutionThing} is conceived for those mental states that represent actions already committed. In the case of processes, the class provides a means for characterizing the realization of the process, the achievement of the related activities, and the commitment of the planned actions. Moreover, to uniquely relate the elements of the process with the ones of its realization, two subproperties of the object-property \textit{drawnBy} are introduced. Specifically, \textit{processDrawnBy} and \textit{procedureDrawnBy} are introduced to connect  the instances of \textit{Process} and \textit{Procedure}, respectively. Hence, it is clear how to trace back to the agent's behaviour responsible for the execution of a process, thus unambiguously identifying actors and actions of arbitrarily complex environments.

\begin{figure}[!htb]
   \centering
   \includegraphics[scale=0.85]{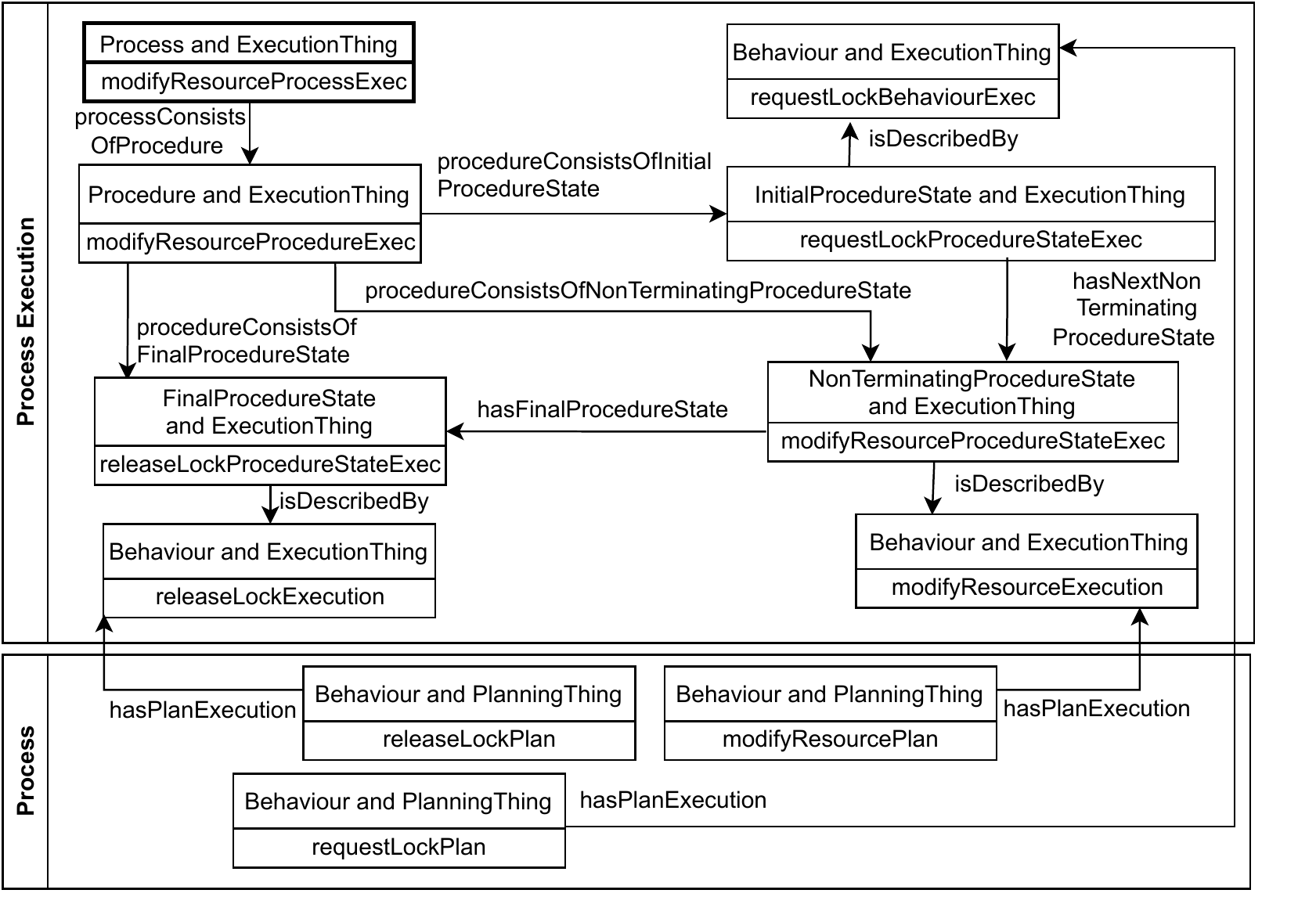}
   \caption{Example of process realization in \ONT{} 2}
    \label{fig:process-impl-example}
\end{figure}

For instance, the process of the example in Fig. \ref{fig:process-example} can be realized as partially depicted in Fig. \ref{fig:process-impl-example} that illustrates the modification process on a resource. The schema retraces the one introduced for the process, where instances of the class \textit{PlanningThing} are suitably replaced by instances of the class \textit{ExecutionThing}. Specifically, the process realization introduces three behaviour executions, one for each plan of the process, namely a) \textit{requestLockBehaviourExec}, representing the execution of the \textit{requestLockPlan} plan; b) \textit{modifyResourceBehaviourExec}, representing the execution of the \textit{modifyResourcePlan} plan; c) \textit{releaseLockBehaviourExec}, representing the execution of the \textit{releaseLockPlan} plan.

Moreover, each behaviour of the process is connected with its analogous of the process realization by means of the object-property \textit{hasPlanExecution}. Finally, each plan execution can be connected with the responsible behaviour by means of the object-property \textit{planExecutionDrawnBy} (subproperty of \textit{executionDrawnBy}, see Section \ref{sec:prelim}).

As stated above, in \ONT{} 2 specific events can be associated with procedure states. In \ONT{} 2 events are semantically represented by instances of the class \textit{Event}, while the procedure state is related to the triggered event by means of the object-property \textit{triggersEvent}. The phenomenon associated with the event is introduced by means of \ONT{} 2 actions. These are in fact sufficiently general and powerful to potentially cover up any type of phenomenon springing from an event that is fully within the vision of a behaviouristic approach.  

\begin{figure}[!htb]
   \centering
   \includegraphics[scale=0.7]{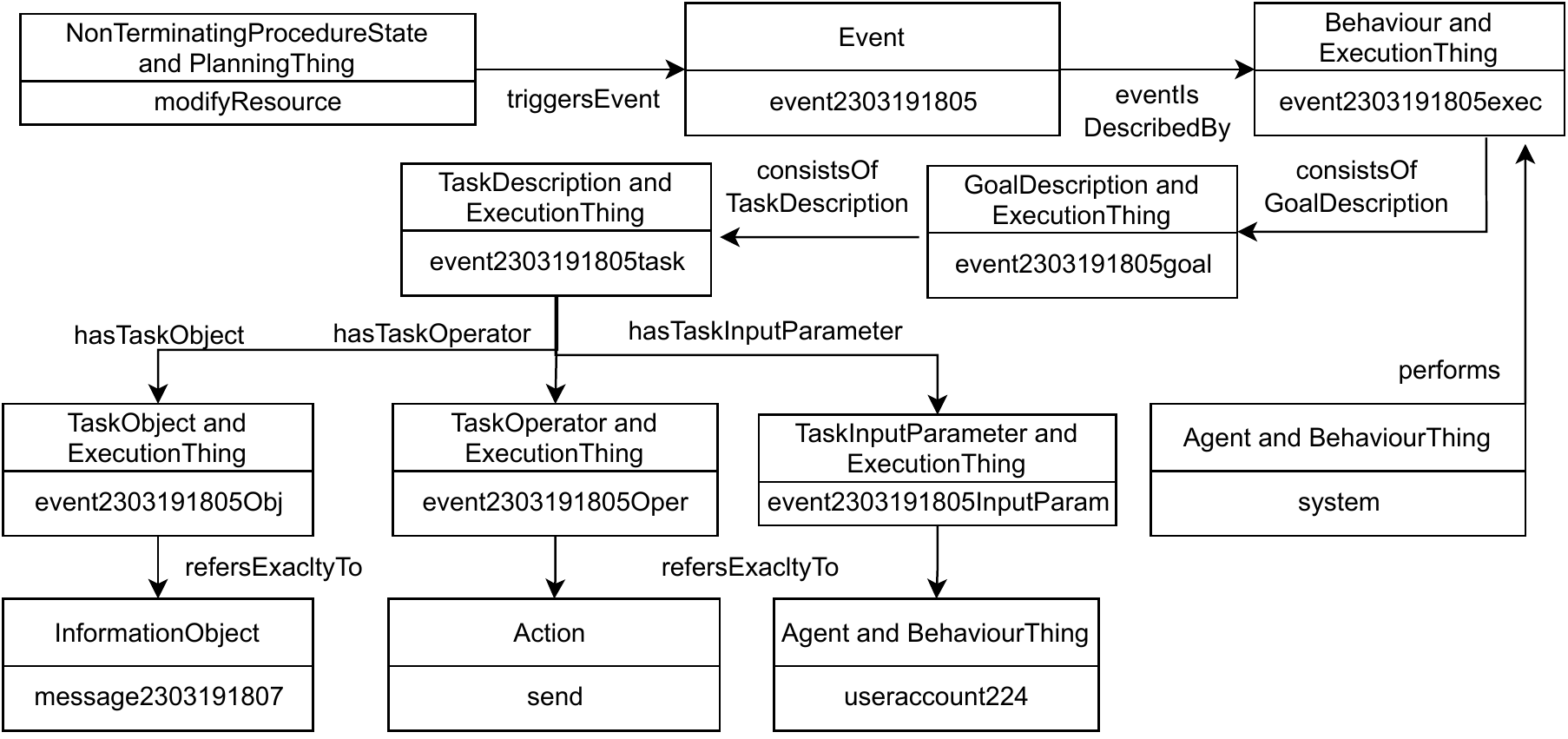}
   \caption{Example of event in \ONT{} 2}
    \label{fig:event-example}
\end{figure}

In the case of the case study, when the system is triggered to send a message to selected users when a resource is modified, for example, because an error occurred, an action such as the one in Fig. \ref{fig:event-example} is introduced. Other information such as the type of event, its duration, or the time window in which it happens, can be additionally specified.

\begin{figure}[!htb]
   \centering
   \includegraphics[scale=0.9]{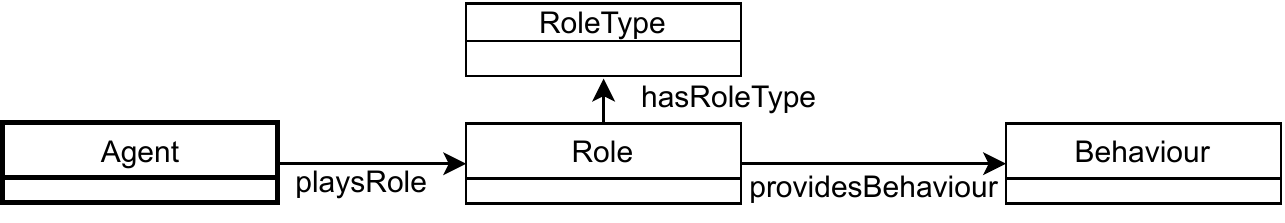}
   \caption{Roles in \ONT{} 2}
    \label{fig:role-schema}
\end{figure}

Finally, it is admissible that agents perform procedures exclusively according to the specific role that they play. Roles permit the separation of the concerns of behaviours intrinsically owned by agents from the ones temporarily at their disposal to execute a process. This implies that the agent is able to perform an action only when such a role is played. Hence, the behaviour responsible for performing those actions is not strictly associated with the agent but with the role. To represent the scenario described above and to model the playing of roles, \ONT{} 2 introduces roles as depicted in Fig. \ref{fig:role-schema}. 

\begin{figure}[!htb]
   \centering
   \includegraphics[scale=0.9]{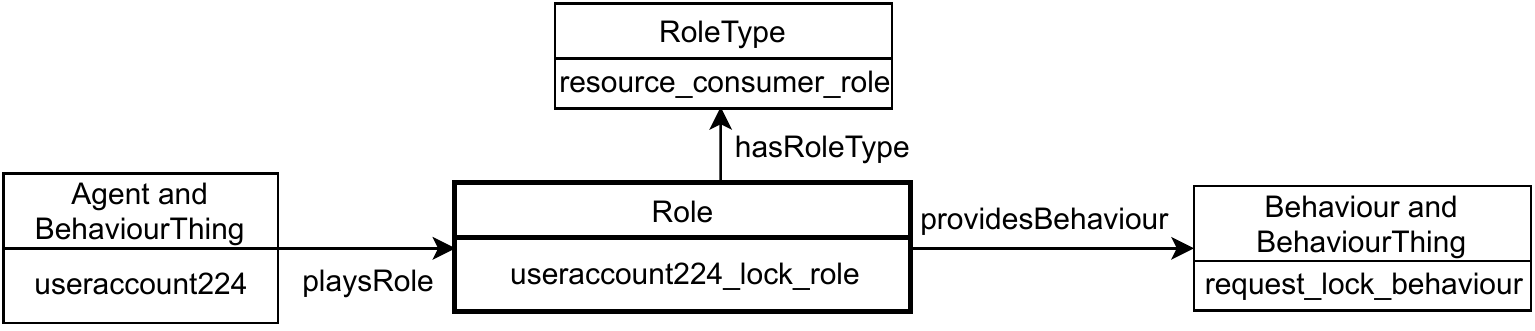}
   \caption{Example of playing roles  in \ONT{} 2}
    \label{fig:role-example}
\end{figure}

In \ONT{} 2 roles, introduced as instances of the class \textit{Role}, provide agents with behaviours by means of the object-property \textit{providesBehaviour}. In its turn, a role is associated with the agent playing it by means of the object-property \textit{playRole}. Whenever agents cease to play a role, the associated behaviours are no longer available. The end of a role is specified by making the instance of \textit{Role} deprecated, hence the instance of the class \textit{DeprecatedThing}. This implies that actions are associated with role-oriented behaviours until the corresponding role is not deprecated. Concerning the case study, an agent playing the role consisting in requesting a lock on a resource is depicted in Fig. \ref{fig:role-example}. 
The role is introduced as a fresh instance of the class \textit{Role}, specifically provided for the agent's representational needs. The role types, introduced as an instance of the class \textit{RoleType}, are defined accordingly to the domain to be described, hence they are out of the \ONT{} 2 scopes: in the case of the case study, a specific role type, called \textit{resource\_consumer\_role}, is introduced to depict the roles dedicated for consuming resources.

\section{Conclusions} \label{ref:conclusions}
This paper introduced an extension of the \ONT{} 2 ontology and, therefore, inherits the behaviouristic approach to semantically represent agents and their commitments through the formalization of their mental states, namely, \emph{behaviour}, \emph{goal} and \emph{task}. In particular, the proposed extension of \ONT{} 2 deals with the modelling of processes and procedures, thereby providing a general although practical way of representing processes and procedures to be tackled by agents through their behaviours. 
Now that such a foundational ontology is available, our semantic representation capabilities are substantially enhanced. In consequence, all application scenarios where specific instructions must be followed have fallen into scope. 

For example, how agents reach a consensus, together with the modelling of their behaviours, is one of the future challenges. An application of \ONT{} 2 is foreseen for the characterisation of security directives, aiming at a structural solution for translating security documents to a mathematically-driven world. The approach targets the NIS 2 directive but other similar directives can be addressed.
In addition, we intend to apply \ONT{} 2 to represent security constraints for cybersecurity threat contexts, in particular for the purpose of semantically representing authentication and confidentiality properties for agents. 

Also, we shall consider how to integrate \ONT{} 2 with the PDDL and with the main frameworks such as JADE \cite{bergenti20}, all with the aim of automatically generating agents and artefacts. Similarly, an integration with \emph{CArtAgO} \cite{Ricci2009}, a framework for building shared computational worlds, appears to be valuable. The horizons of the semantic representation of agents have substantially expanded but retain the vast potential for yet more expansion in the near future.

\section*{Acknowledgments}
Gianpietro Castiglione acknowledges a studentship by Intrapresa S.r.l. and Italian ``Ministero dell’Università e della Ricerca'' (D.M. n. 352/2022).

\bibliography{cite}
\end{document}